\documentclass{article}
\usepackage{ijcai24}

\usepackage{times}
\usepackage{soul}
\usepackage{url}
\usepackage[hidelinks]{hyperref}
\usepackage[utf8]{inputenc}
\usepackage[small]{caption}
\usepackage{graphicx}
\usepackage{amsmath}
\usepackage{amsthm}
\usepackage{booktabs}
\usepackage{algorithm}
\usepackage{algorithmic}
\usepackage[switch]{lineno}
\usepackage{bbding}

\usepackage{subfig}
\usepackage{pifont}
\usepackage{enumitem}
\setlist{leftmargin=5.5mm}

\usepackage{multicol}
\usepackage{multirow}

\title{WikiTableEdit: A Benchmark for Table Editing by Natural Language Instruction}

\author{
Zheng Li$^1$
\and
Xiang Chen$^2$\and
Xiaojun Wan$^3$\\
\affiliations
$^1$Peking University\\
$^2$Peking University\\
$^3$Peking University\\
\emails
\{lizheng2001, caspar, wanxiaojun\}@pku.edu.cn
}

\begin{document}

\maketitle

\begin{abstract}
Tabular data, as a crucial form of data representation, exists in diverse formats on the Web. When confronted with complex and irregular tables, manual modification becomes a laborious task. This paper investigates the performance of Large Language Models (LLMs) in the context of table editing tasks. Existing research mainly focuses on regular-shaped tables, wherein instructions are used to generate code in SQL, Python, or Excel Office-script for manipulating the tables. Nevertheless, editing tables with \textbf{irregular} structures, particularly those containing merged cells spanning multiple rows, poses a challenge when using code. To address this, we introduce the WikiTableEdit dataset. Leveraging 26,531 tables from the WikiSQL dataset, we automatically generate natural language instructions for six distinct basic operations and the corresponding outcomes, resulting in over 200,000 instances. Subsequently, we evaluate several representative large language models on the WikiTableEdit dataset to demonstrate the challenge of this task. The dataset will be released to the community to promote related researches\footnote{Some dataset samples are available at \url{https://anonymous.4open.science/r/WikiTableEdit-ECEC} (anonymous link).}. 
\end{abstract}

\section{Introduction}

Tabular data plays a crucial role in real-life scenarios, widely employed for organizing, storing, and analyzing information. Table editing is a common way of data manipulation by modifying existing tables to meet specific requirements, which is usually a laborious task.   Previous research on table editing has predominantly focused on implementing table edits through programming languages, generating code in SQL, Python, or Excel Office-script for table manipulation.

In this study, we aim to construct a high-quality table editing dataset and evaluate large language models' capabilities on this task. Previous approaches to editing through code generation have imposed constraints on the form of source and target tables, requiring tables to adhere to regular shapes and necessitating the execution of programming languages for table edits, like InstructExcel~\cite{payan2023instructexcel}. To achieve a visually concise and comprehensible representation, real-world tables often deviate from regular shapes by merging adjacent cells with identical content. Additionally, executing code can be challenging for non-professional users. Given the impressive capabilities demonstrated by LLMs in the field of text editing, we explore the direct treatment of table editing as a distinct form of long-text editing task. This task removes constraints on the form of tables, allowing for both regular database tables and irregular web tables. Moreover, it eliminates the need for non-professionals to engage in debugging or code execution steps, enabling them to obtain the modified table directly.

We introduce \textbf{WikiTableEdit}, a benchmark for both regular and irregular table editing by natural language instruction. WikiSQL~\cite{zhong2017seq2sql} is a dataset that consists of 26,531 tables extracted from Wikipedia, with associated natural language questions, SQL queries, and the corresponding execution results. In our task, we utilize tables sourced from WikiSQL as the original tables. We employ a programmatic method to automatically generate triplets consisting of natural language instruction, source table, and target table under various operations. Our dataset encompasses editing operations on both regular and irregular tables.

Furthermore, we conduct experiments and analyses on the WikiTableEdit dataset to substantiate its characteristics and challenges. We conduct comparisons of multiple LLMs before and after fine-tuning and also evaluate the performance of GPT-3.5 on this task. To better quantify the effectiveness of table generation, we design a new metric, the table edit distance. We believe that the WikiTableEdit dataset can facilitate future research in delving deeper into the task of table editing. 

Our primary contributions can be summarized as follows:
\begin{itemize}

\item We introduce the task of table editing and construct a high-quality dataset \textbf{WikiTableEdit} to support this endeavor. Furthermore, we design a metric Table Edit Distance (TED) to evaluate the differences and similarities between source and target tables.

\item We conduct extensive experiments and analyses on the dataset with popular LLMs and manually analyze the results. We find that swap and reorder operations on tables are the most challenging ones for the table editing task. 

\end{itemize}
\begin{figure*}[!h]

\includegraphics[width=7in]{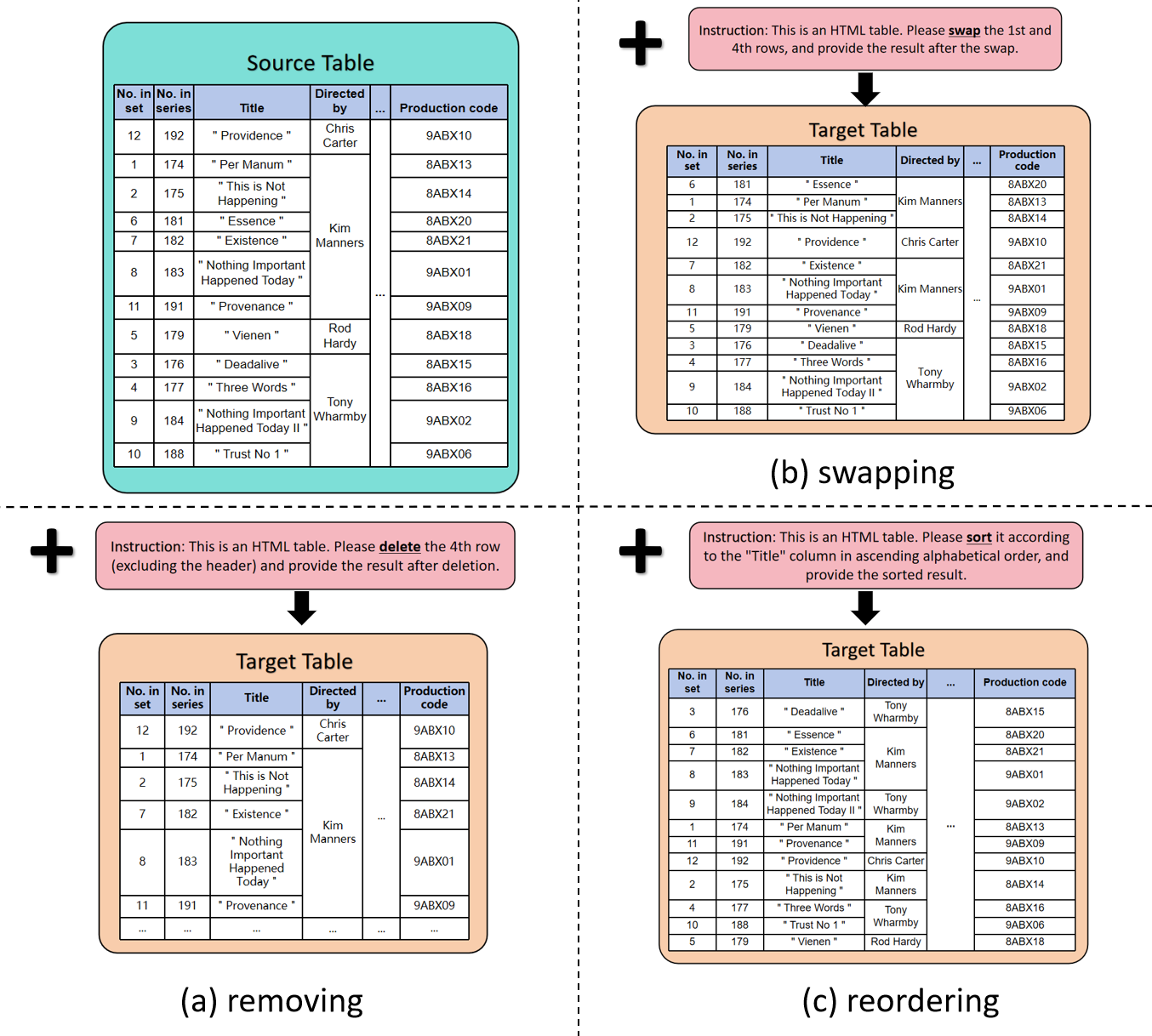}

\caption{Illustration for the table editing task.}
\label{fig:dataset-eg}
\end{figure*}

\section{Task Overview}

In this study, we aim to explore the capability of LLMs in directly editing tables. Thus, each data instance in the table editing task consists of a triplet denoted as (Instruction, Source table, Target table)\footnote{More specific examples of the dataset can be found in the Appendix.}, where:
\begin{enumerate}
    \item \textbf{Instruction}: a natural language instruction indicating how to modify the source table into the target table with a specific operation.
    \item \textbf{Source Table}: the original table before editing.
    \item \textbf{Target Table}: the revised table after editing. 
\end{enumerate}
Both the source table and target table may exhibit either regular or irregular structures. Figure~\ref{fig:dataset-eg} gives examples of the table editing task.

\section{Dataset Construction}

Considering the formidable text editing capabilities of LLMs, our dataset places a greater emphasis on editing the structure of tables rather than the textual content within them. We select six commonly-employed fundamental operations for our dataset: (1) \textbf{Adding} a new row or column, (2) \textbf{Removing} a row or column, (3) \textbf{Swapping} two rows, (4) \textbf{Reordering} based on a certain column, (5) \textbf{Merging} adjacent cells with identical values, and (6) \textbf{Splitting} the merged cells.

We take the regular tables from the WikiSQL dataset~\cite{zhong2017seq2sql} as the most primitive data and develop a program to automatically generate triplets (Instruction, Source table, Target table) for the six types of editing operations. Our dataset can be divided into two main parts: \textbf{regular table editing} and \textbf{irregular table editing}. For regular table editing, we can directly use the primitive table as the source table, requiring only the generation of instructions and the edited result. For irregular table editing, it is necessary to additionally generate the irregular source table before editing.

\subsection{Automatic Generation}

In this section, we will offer a comprehensive explanation of how we automatically generate data for the six editing operations, corresponding to both regular and irregular table editing scenarios. To facilitate the ingestion of tables, especially irregular tables, by LLMs, we represent the tables in HTML format.

\begin{itemize}

\item \textbf{Adding} a new row or column. Generating a single row or column of new data that is contextually appropriate can be challenging. Thus, we designate the original table as the target table. A random selection of a row or column from the table becomes the new data, and the table resulting from removing that specific row or column is considered the source table. Instructions will provide details about the new data to be added and specify its designated location. We design natural language templates to generate the instructions. 

\item \textbf{Removing} a row or column. The original table serves as the source table. A random selection of a row or column from the table is identified as the data to be deleted. The resulting table, after removing that specific row or column, is considered the target table. Instructions will include details about the location of the data to be deleted.

\item \textbf{Swapping} two rows. The original table serves as the source table. Two rows are randomly selected as the data to be swapped. The resulting table, after performing the swap operation, is considered the target table. Instructions will include details about the location of the rows to be swapped.

\item \textbf{Reordering} based on a certain column. The original table is employed as the source table. One column is randomly selected to serve as the key value for sorting. The resulting table, after sorting based on the chosen key column, is considered the target table. Instructions will include information about the selected key column.

\item \textbf{Merging} adjacent cells with identical values. The original table is used as the source table. We iterate through each column in the table and attempt to perform merging across rows. In tables where attributes for different entities differ, it is not possible to execute merging operations and generate corresponding data. For tables where merging is feasible, instructions will provide information about the columns that can be merged.

\item \textbf{Splitting} the merged cells. Similar to the adding operation, the original table is designated as the target table, and the table resulting from performing the merge operation is considered the source table. For these types of operations, the instruction does not provide additional information; it simply requires the LLMs to split all merged cells.

\end{itemize}

For the irregular table editing, after performing the aforementioned operations on the source table and the target table, we need to conduct cell merging operations on both tables.

\subsection{Dataset Analysis}
\begin{table}[t!]
    \centering
    \begin{tabular}{l|cccc}
    \toprule
    \multirow{2}{*}{Operations} & \multicolumn{2}{c}{Regular} & \multicolumn{2}{c}{Irregular}   \\
        & Row & Column & Row & Column \\
    \midrule
    Adding & \Checkmark & \Checkmark & \Checkmark & \XSolidBrush \\
    Removing & \Checkmark & \Checkmark & \Checkmark & \XSolidBrush \\
    Swapping & \Checkmark & \XSolidBrush  & \Checkmark & \XSolidBrush \\
    Reordering & \Checkmark & \XSolidBrush  & \Checkmark & \XSolidBrush \\
    Merging & \XSolidBrush & \XSolidBrush & \XSolidBrush & \Checkmark \\
    Splitting & \XSolidBrush & \XSolidBrush & \XSolidBrush & \Checkmark \\
    \bottomrule
    \end{tabular}
    
    \caption{The combination of operation types, table types, and operation objects (Row or Column) in WikiTableEdit.}
    \label{tab: operation type matrix}
\end{table}

Due to the inherent involvement of irregular tables in merge and split operations, only the first four operations will have both regular and irregular table versions. Additionally, in the WikiSQL dataset, each column represents a property, and each row represents an entity. Therefore, the irregular tables we generate often involve merging across rows rather than across columns. In such cases, operations on columns in irregular tables are almost identical to those in regular tables. Hence, for irregular tables, we only generate versions involving operations on rows. 

As shown in the Table~\ref{tab: operation type matrix}, an original table in WikiSQL dataset can generate up to twelve table editing data instances. We utilized 24,241 tables from the training set of WikiSQL to generate 194,996 data instances, which form our training set. Additionally, 2716 tables from the development set were used to generate 28,706 instances, constituting our test set. Considering the current limitations on the context length of LLMs, we have filtered out some excessively long data instances.


\subsection{Examples of WikiTableEdit}

Figure~\ref{fig:dataset-eg} illustrates some examples of WikiTableEdit. Considering that editing regular tables is more straightforward to comprehend, we showcase three editing operations on an irregular table, i.e., removing, swapping and reordering.

To conserve storage space, details about merged cells in a table are not stored for every row but solely in the first row of the merged cell. For instance, in the ``Directed by'' column of the source table in Figure~\ref{fig:dataset-eg}, ``Kim Manners'' will only appear in the second row along with the associated information ``rowspan=6''. Hence, when a model edits irregular tables, it must identify information about merged cells. The complexity of editing irregular tables significantly increases as the model needs to calculate the new rowspan for merged cells post-operation and update the corresponding positions. 

As illustrated in Figure~\ref{fig:dataset-eg}(b), when swapping the first and fourth rows, it becomes imperative to retrieve information about the merged cell ``Kim Manners'' from the second row before executing the swap operation. Following the completion of the swap, the initial merged cell ``Kim Manners'' is split into two smaller merged cells, necessitating updates to the corresponding information in the first and fifth rows of the target table.

\begin{table*}[!h]
\centering
\small
\setlength{\tabcolsep}{3mm}{
\begin{tabular}{l|ccc|cc}
\toprule
\textbf{Model}           & \textbf{EM} & \textbf{SARI} & \textbf{TED} & \textbf{BLEU} & \textbf{ROUGE-L}\\
 \midrule
 LLaMA2-7B & 0 & 12.02 & 15.49 & 1.32 & 28.39  \\
 LLaMA2-7B (few-shot) &  0 & 30.94  & 61.16 & 24.29  & 64.23   \\

 \midrule
 ChatGLM3-6B & 0 & 23.30  & 33.06 &  17.83 & 53.35   \\
 ChatGLM3-6B (few-shot) & 1.13  & 24.85  & 36.31 & 14.92 & 49.23   \\
\midrule
 Falcon-7B & 0  & 3.74  & 0.34 &  0.18 & 9.02   \\
 Falcon-7B (few-shot) & 0  & 6.05  & 0.28 & 0.19  & 15.27   \\
\midrule
 GPT-3.5-turbo & 6.34 & 54.21  & 53.81 & 13.72  & 65.74   \\
 GPT-3.5-turbo (few-shot) &  5.32 & 28.99  & 32.92 & 8.96  & 33.78   \\
\bottomrule

\end{tabular}}
\caption{Performance of different models on the WikiTableEdit test set.}
\label{tab:intent}
\end{table*}
\section{Experiments}

To investigate the capabilities of LLMs in table editing tasks, we conducted experiments on LLaMA2-7B~\cite{touvron2023llama}, ChatGLM3-6B~\cite{du2022glm}, and Falcon-7B~\cite{falcon40b}. We evaluated the results of these models under zero-shot, few-shot, and supervised fine-tuning scenarios, respectively\footnote{Due to limited computing resources, we were not able to fine-tune larger LLMs, and thus mainly report results of LLMs with less than 10M parameters.}. Additionally, we performed zero-shot and few-shot evaluations for GPT-3.5-turbo~\cite{openai2023gpt4}.

\subsection{Evaluation Metrics}

We utilized the following popular metrics to assess the quality of the model-generated results:
\begin{itemize}
    \item \textbf{Exact Match (EM)}: The ratio of correctly edited instances. An editing operation is considered correct if and only if the result is consistent with the reference.
    \item \textbf{SARI}~\cite{SARI}: The average F1 score for keep, add and delete operations.
    \item \textbf{BLEU}~\cite{BLEU}: An important metric in the machine translation domain, its reliability is limited due to significant redundancy between inputs and outputs. Therefore, it should be considered only as a reference.
    \item \textbf{ROUGE}~\cite{lin2004rouge}: An important metric in the text summarization domain, Due to reasons similar to those impacting BLEU, this metric is also considered unreliable and should be used solely for reference purposes.
\end{itemize}

Considering the specialty of tables, we designed a custom evaluation metric \textbf{Table Edit Distance (TED)} that considers both the minimum edit distance of table structures and the minimum edit distance of table content.

In table editing tasks, when tables are presented in HTML format, the content length is often extensive, and modifications usually impact only a small section of it. When employing metrics such as BLEU and ROUGE to directly measure the similarity between the source and target tables, even a straightforward copy of the source table can result in high scores. Hence, an appropriate evaluation metric should enable a direct comparison of the similarity between the editing operations performed from source to reference and the editing operations undertaken from source to the model's predicted output. For example, SARI is an excellent choice for this purpose. Moreover, given the relatively low subjectivity in table editing, where individuals tend to generate consistent results for identical instructions, Exact Match emerges as a valuable metric.

While each of the previously mentioned metrics possesses its strengths and weaknesses, none explicitly accounts for alterations in table structure. To fill this gap, we have introduced Table Edit Distance (TED), a metric that concurrently assesses changes in both table structure and content.

The TED score is composed of two key components: content and structure. To simplify the comparison of content differences, we initially decompose all merged cells in irregular tables, converting them into regular tables. Subsequently, treating individual cells as the smallest editing units, we regard additions, deletions, and modifications of cell content as basic operations. The minimum edit distance for content between two tables is then calculated.

\begin{figure}[!h]
\centering
\includegraphics[width=3.5in]{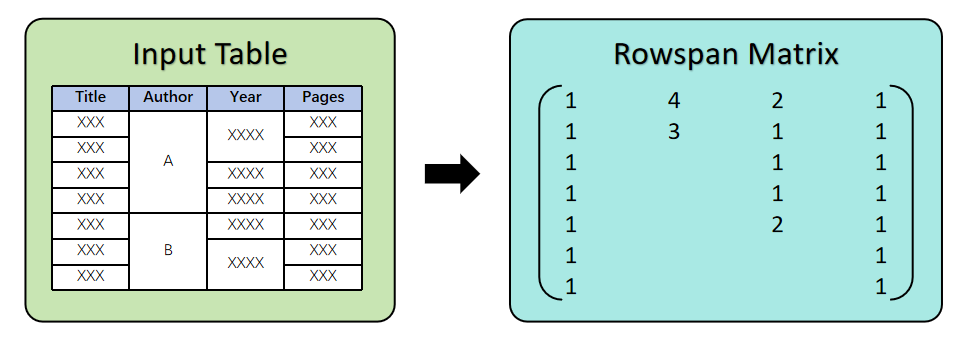}
\caption{An example for rowspan matrix extraction.}
\label{fig:rowspanmatrix}
\end{figure}

On the structural side, we also treat individual cells as the smallest units. To differentiate between irregular and regular tables, we refrain from converting irregular tables into regular ones.  As shown in Figure~\ref{fig:rowspanmatrix}, for each table we initially extract its rowspan matrix. Then, considering additions, deletions, and splits as fundamental operations, we determine the minimum edit distance between two rowspan matrices, representing the structural edit distance between the structures of the two tables.

Finally, we add the structural distance and content distance to obtain the overall distance. Given that the overall distance between source table and reference table in this test is typically much smaller than 100, to offer a more intuitive grasp of the disparities between the two tables, we subtract the overall distance from 100, resulting in the final TED score.

In addition to the above automatic evaluation metrics, we randomly sample a portion of results, and manually verify their correctness.

\subsection{Experiment Details}

For zero-shot evaluation, we concatenate the instruction and source table directly as the model input, testing the LLMs without any supervised fine-tuning.

For few-shot testing, given the model's context length limitations, we include only one example in the input, ensuring that the total length does not exceed the specified limit after adding the example. The edit type of the example aligns with the desired model completion type. For instance, if the test example involves reordering irregular tables, the provided example will also relate to reordering irregular tables. The input will be formatted as follows: \textit{"Please generate the corresponding target table following the example below. Input: }\verb|{example instruction + source table}| \textit{Output:} \verb|{example target table}| \textit{Input:} \verb|{test instruction + source table}| \textit{Output:"} 

For supervised fine-tuning, we initially use the 194,996 examples from the training set for fine-tuning the language models. After fine-tuning, similar to zero-shot evaluation, we concatenate the instruction and source table, using it as the model input without providing any additional examples. 


\section{Results}

\begin{table*}[!h]
\centering
\small
\setlength{\tabcolsep}{3mm}{
\begin{tabular}{lccccc}
\toprule
\textbf{Operation}           & \textbf{EM} & \textbf{SARI} & \textbf{TED} & \textbf{BLEU} & \textbf{ROUGE-L}\\
 \midrule
 Add-row & 0 & 15.41 & 20.95 & 1.59 & 34.59  \\
 Add-col & 0  &  8.42 & 9.30 & 0.60  & 17.90    \\
 Add-irreg & 0 & 15.70  & 19.44 &  1.68 & 35.07    \\
 Remove-row & 0 & 15.55 & 22.00 & 2.38  & 36.09   \\
 Remove-col & 0 & 24.00  & 40.83 &  1.45 & 47.77   \\
 Remove-irreg & 0  & 17.52  & 29.08 & 1.64  & 41.94   \\
 Swap & 0  &  6.39 & 1.82 & 1.39  & 17.38   \\
 Swap-irreg & 0  & 6.51  & 2.48 &  1.05 & 17.35   \\
 Reorder & 0  & 5.67  & 6.12 & 0.88  & 17.46  \\
 Reorder-irreg & 0  &  6.10 & 6.41 & 0.84  & 17.24   \\
 Merge &  0 & 13.13  & 17.56 & 1.12  & 32.62   \\
 Split &  0 & 9.40  & 8.75 & 0.99  & 24.01   \\
\bottomrule
\end{tabular}}
\caption{The zero-shot performance of LLaMA2-7B on different operations in the test set. In the Operation Column, ``row" denotes row operations, ``col" denotes column operations, and ``irreg" indicates operations targeted at irregular tables. }
\label{tab:ops}
\end{table*}

\begin{table*}[!h]
\centering
\small
\setlength{\tabcolsep}{3mm}{
\begin{tabular}{l|ccccc|ccccc}
\toprule
\multirow{3}{*}{\textbf{Model}} & \multicolumn{5}{c|}{\multirow{2}{*}{{\textbf{Regular Table}}}} & \multicolumn{5}{c}{\multirow{2}{*}{{\textbf{Irregular Table}}}} \\
&  & & & & & & & & &   \\
& \textbf{EM} & \textbf{SARI} & \textbf{TED} & \textbf{BLEU} & \textbf{ROUGE-L}& \textbf{EM} & \textbf{SARI} & \textbf{TED} & \textbf{BLEU} & \textbf{ROUGE-L}\\
 \midrule
 LLaMA2-7B            & 0 & 12.60 & 16.92 & 1.41 & 28.71    & 0 & 11.40 & 13.96 & 1.22 & 26.91  \\
 ChatGLM3-6B          & 0 & 32.95 & 52.72 & 33.11 & 68.04     & 0 & 15.30 & 16.76 & 5.17 &  41.17 \\
 Falcon-7B            & 0 & 4.24 & 0.61 &  0.33 &  13.08   & 0 & 3.32 & 0.12 & 0.06 &  5.66 \\
 GPT-3.5-turbo & 11.08 & 60.95 & 59.42 & 17.43 & 69.23     & 1.27 & 46.99 & 47.81 & 9.76 & 62.00 \\
\bottomrule
\end{tabular}}
\caption{The model's performance under zero-shot conditions is compared in terms of editing regular and irregular tables in WikiTableEdit.}
\label{tab:ir_vs_reg}
\end{table*}
The results of all models are presented in Table ~\ref{tab:intent}. 
GPT-3.5-turbo demonstrates outstanding performance, surpassing other LLMs by a significant margin in its ability to handle table editing tasks, especially in zero-short settings. It significantly outperforms other models across all metrics. However, the performance of GPT-3.5-turbo is still far from satisfactory, as noted by the very low EM score. 

Few-shot prompting yields minimal improvements in table editing tasks. Across various metrics, providing examples does not result in a significant enhancement in any model's performance; at times, there is even a slight decrease. We speculate that this might be due to two potential reasons. Firstly, the current model's context length limitations are relatively short for table editing tasks. With substantial overlap between input and output in the examples, the model might struggle to capture changes in the table and establish correspondences between changes and instructions. Secondly, the number of examples provided may be insufficient. Constrained by the model's context length, it is challenging to include more examples in the instruction.

Supervised fine-tuning brings about a significant improvement. After the fine-tuning process, all models show nearly doubled results across various metrics. LLaMA2-7B and ChatGLM3-6B achieve complete consistency with the reference answers on approximately a quarter of the test data. Although Falcon-7B still performs relatively poorly after fine-tuning, there is a substantial improvement compared to its performance before fine-tuning. Due to cost considerations, we did not conduct fine-tuning on GPT-3.5-turbo. Noting that supervised fine-tuning is not our primary focus, we have placed the results in the Appendix.

\subsection{Different Editing Operations}

To further investigate the model's performance on different types of operations, we separately calculated various metrics for LLaMA2-7B on the test set under zero-shot conditions, and the results are shown in Table~\ref{tab:ops}. 

The overall performance of LLaMA2-7B is subpar. Relatively speaking, the model performs best on deletion operations, followed closely by addition operations. These two are the most basic model operations, requiring the model to have a fundamental understanding of table data. Next are merge and split operations, which demand the model's ability to comprehend and express irregular tables. The most challenging operations are swap and reorder operations, which can be considered as multiple instances of deletion and addition operations. Reordering not only requires the model to smoothly handle addition and deletion tasks but also places certain demands on the model's mathematical abilities.

\subsection{Regular vs. Irregular Table Editing}

This section presents a comparison of the models' abilities in editing regular and irregular tables under zero-shot conditions, with the results summarized in Table~\ref{tab:ir_vs_reg}.

It is worth noting that the LLaMA2-7B and Falcon-7B exhibit no significant difference in performance between editing regular and irregular tables. We attribute this to the limited capability of these models in handling table editing tasks. The inherent difficulty variations among various operations obscure the distinctions between regular and irregular table editing, posing a challenge in observing a clear difficulty gap between the two.

Notably, as the models' proficiency in handling table editing improves, the performance gap between regular and irregular tables becomes more evident. In the case of ChatGLM3-6B, all metrics, except EM, exhibit a decrease ranging from 21\% to 82\%. Meanwhile, GPT-3.5-turbo experiences a direct performance degradation in EM to just 11\% of the original.

\subsection{Manually Check}

For each model, we randomly examined a portion of the test results under zero-shot conditions and attempted to summarize their performance as follows.  

Falcon-7B struggles to comprehend the intent of instructions, often producing either irrelevant content or simplifying the instructions and presenting them as results.

LLaMA2-7B performs slightly better than Falcon-7B. In almost half of the cases, it produces incomplete tables. In the remaining half, it either issues an apology or generates a segment of relevant text.

ChatGLM3-6B demonstrates some degree of table understanding, albeit very limited. In the majority of cases, its output contains a table, even if the table may be incomplete. Occasionally, it manages to complete a portion of the instructions, while at other times, it merely repeats the source table and instructions.

GPT-3.5-turbo exhibits the best performance. When faced with small and regular tables, it can provide correct answers. However, as the tables grow in size, it becomes more prone to operational errors, such as misidentifying the location of the row to be manipulated. Regarding the editing of irregular tables, it struggles to handle the table structure and often generates tables with peculiar shapes. Additionally, in our random inspections, we observed a phenomenon where, in nearly one-third of cases, it apologizes and refuses to answer. In a few instances, after stating its refusal to answer, it proceeds to provide a modified table.

\section{Related Works}

Previous research in table-related studies has predominantly concentrated on TableQA and Data2text. TableQA involves investigating methods for generating accurate responses to provided tables and natural language queries. Prominent datasets within this domain include WebQuestions~\cite{webq}, WikiTableQuestions~\cite{wtq}, WikiSQL~\cite{zhong2017seq2sql}, and Text2Analysis~\cite{he2023text2analysis}. Data2text predominantly explores the creation of suitable descriptions or summaries derived from tables. Noteworthy datasets within this domain encompass WeatherGov~\cite{mei2016talk}, RotoWire~\cite{wiseman2017challenges}, and WikiTableT~\cite{chen2021wikitablet}. 

Nevertheless, these studies do not center around table editing; instead, their emphasis lies in delving into the information embedded within the table content. Recently, endeavors have emerged to tackle table editing through code. One example is InstructExcel~\cite{payan2023instructexcel}, which leverages Microsoft's Office Script to edit Excel files. Nevertheless, manipulating tables via code demands users employing the models to possess a certain level of coding proficiency for debugging, which may pose an inconvenience. To mitigate these challenges, we propose WikiTableEdit, for directly editing tables by natural language instruction. 


\section{Conclusion}

In this paper, we introduce WikiTableEdit, a benchmark for table editing guided by natural language instructions. WikiTableEdit encompasses both regular table editing and irregular table editing, covering six different types of operations. It includes 194,996 training data instances and 28,706 testing data instances. To explore the table editing capabilities of LLMs, we devised a new automated metric, TED, and conducted extensive and diverse experiments. The current results indicate that existing models still have a long way to go in terms of table editing. We believe that WikiTableEdit, along with the metrics and tasks we designed, can contribute to helping models truly grasp the diverse forms of tabular data.

\section{Limitations}

WikiTableEdit is a programmatically generated dataset with a relatively uniform instruction format, where each editing operation has only one form of instruction. Although we generated data in English, the same process can be applied to generate data in other languages after collecting corresponding language tables. Additionally, due to GPU resource limitations, we were unable to test the performance of larger-scale open-source models on this task




\bibliographystyle{named}
\bibliography{ijcai24}

\begin{thebibliography}{}

\bibitem[\protect\citeauthoryear{Almazrouei \bgroup \em et al.\egroup }{2023}]{falcon40b}
Ebtesam Almazrouei, Hamza Alobeidli, Abdulaziz Alshamsi, Alessandro Cappelli, Ruxandra Cojocaru, Merouane Debbah, Etienne Goffinet, Daniel Heslow, Julien Launay, Quentin Malartic, Badreddine Noune, Baptiste Pannier, and Guilherme Penedo.
\newblock {Falcon-40B}: an open large language model with state-of-the-art performance.
\newblock 2023.

\bibitem[\protect\citeauthoryear{Berant \bgroup \em et al.\egroup }{2013}]{webq}
Jonathan Berant, Andrew Chou, Roy Frostig, and Percy Liang.
\newblock Semantic parsing on freebase from question-answer pairs.
\newblock In {\em Proceedings of the 2013 conference on empirical methods in natural language processing}, pages 1533--1544, 2013.

\bibitem[\protect\citeauthoryear{Chen \bgroup \em et al.\egroup }{2021}]{chen2021wikitablet}
Mingda Chen, Sam Wiseman, and Kevin Gimpel.
\newblock Wikitablet: A large-scale data-to-text dataset for generating wikipedia article sections, 2021.

\bibitem[\protect\citeauthoryear{Du \bgroup \em et al.\egroup }{2022}]{du2022glm}
Zhengxiao Du, Yujie Qian, Xiao Liu, Ming Ding, Jiezhong Qiu, Zhilin Yang, and Jie Tang.
\newblock Glm: General language model pretraining with autoregressive blank infilling.
\newblock In {\em Proceedings of the 60th Annual Meeting of the Association for Computational Linguistics (Volume 1: Long Papers)}, pages 320--335, 2022.

\bibitem[\protect\citeauthoryear{He \bgroup \em et al.\egroup }{2023}]{he2023text2analysis}
Xinyi He, Mengyu Zhou, Xinrun Xu, Xiaojun Ma, Rui Ding, Lun Du, Yan Gao, Ran Jia, Xu~Chen, Shi Han, Zejian Yuan, and Dongmei Zhang.
\newblock Text2analysis: A benchmark of table question answering with advanced data analysis and unclear queries, 2023.

\bibitem[\protect\citeauthoryear{Lin}{2004}]{lin2004rouge}
Chin-Yew Lin.
\newblock Rouge: A package for automatic evaluation of summaries.
\newblock In {\em Text summarization branches out}, pages 74--81, 2004.

\bibitem[\protect\citeauthoryear{Mei \bgroup \em et al.\egroup }{2016}]{mei2016talk}
Hongyuan Mei, Mohit Bansal, and Matthew~R. Walter.
\newblock What to talk about and how? selective generation using lstms with coarse-to-fine alignment, 2016.

\bibitem[\protect\citeauthoryear{OpenAI \bgroup \em et al.\egroup }{2023}]{openai2023gpt4}
OpenAI, :, Josh Achiam, Steven Adler, Sandhini Agarwal, Lama Ahmad, Ilge Akkaya, et~al.
\newblock Gpt-4 technical report, 2023.

\bibitem[\protect\citeauthoryear{Papineni \bgroup \em et al.\egroup }{2002}]{BLEU}
Kishore Papineni, Salim Roukos, Todd Ward, and Wei-Jing Zhu.
\newblock Bleu: A method for automatic evaluation of machine translation.
\newblock In {\em Proceedings of the 40th Annual Meeting on Association for Computational Linguistics}, ACL '02, page 311–318, USA, 2002. Association for Computational Linguistics.

\bibitem[\protect\citeauthoryear{Pasupat and Liang}{2015}]{wtq}
Panupong Pasupat and Percy Liang.
\newblock Compositional semantic parsing on semi-structured tables.
\newblock {\em CoRR}, abs/1508.00305, 2015.

\bibitem[\protect\citeauthoryear{Payan \bgroup \em et al.\egroup }{2023}]{payan2023instructexcel}
Justin Payan, Swaroop Mishra, Mukul Singh, Carina Negreanu, Christian Poelitz, Chitta Baral, Subhro Roy, Rasika Chakravarthy, Benjamin~Van Durme, and Elnaz Nouri.
\newblock Instructexcel: A benchmark for natural language instruction in excel, 2023.

\bibitem[\protect\citeauthoryear{Touvron \bgroup \em et al.\egroup }{2023}]{touvron2023llama}
Hugo Touvron, Louis Martin, Kevin Stone, Peter Albert, Amjad Almahairi, et~al.
\newblock Llama 2: Open foundation and fine-tuned chat models, 2023.

\bibitem[\protect\citeauthoryear{Wiseman \bgroup \em et al.\egroup }{2017}]{wiseman2017challenges}
Sam Wiseman, Stuart~M. Shieber, and Alexander~M. Rush.
\newblock Challenges in data-to-document generation, 2017.

\bibitem[\protect\citeauthoryear{Xu \bgroup \em et al.\egroup }{2016}]{SARI}
Wei Xu, Courtney Napoles, Ellie Pavlick, Quan~Ze Chen, and Chris Callison-Burch.
\newblock Optimizing statistical machine translation for text simplification.
\newblock {\em Transactions of the Association for Computational Linguistics}, 4:401--415, 2016.

\bibitem[\protect\citeauthoryear{Zhong \bgroup \em et al.\egroup }{2017}]{zhong2017seq2sql}
Victor Zhong, Caiming Xiong, and Richard Socher.
\newblock Seq2sql: Generating structured queries from natural language using reinforcement learning, 2017.

\end{thebibliography}

\clearpage
\appendix
\section{Detailed Examples of WikiTableEdit}

\begin{figure*}[!t]
\centering
\subfloat[Adding a new row to the regular table.]{
		\includegraphics[scale=0.33]{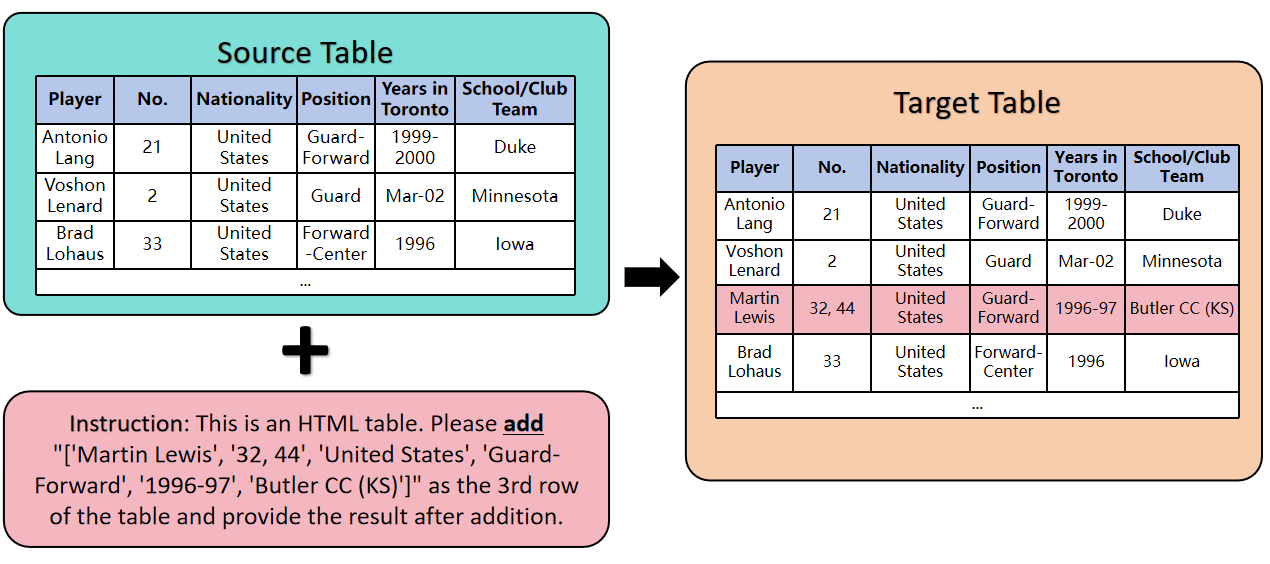}}
\subfloat[Adding a new column to the regular table.]{
		\includegraphics[scale=0.33]{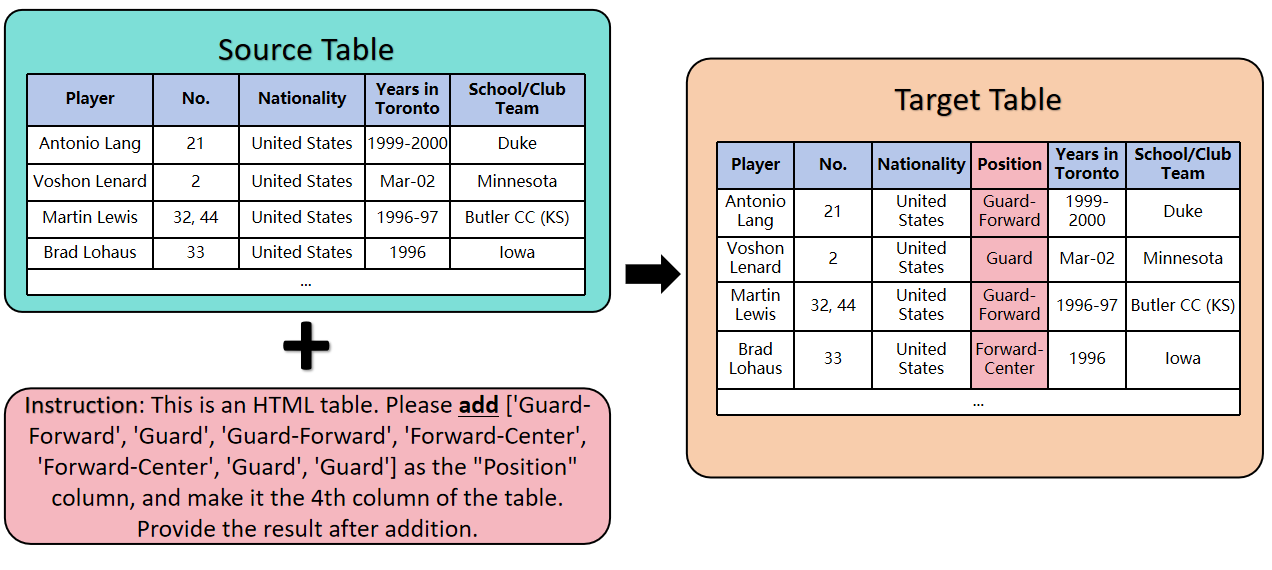}}
\\
\subfloat[Adding a new row to the irregular table.]{
		\includegraphics[scale=0.33]{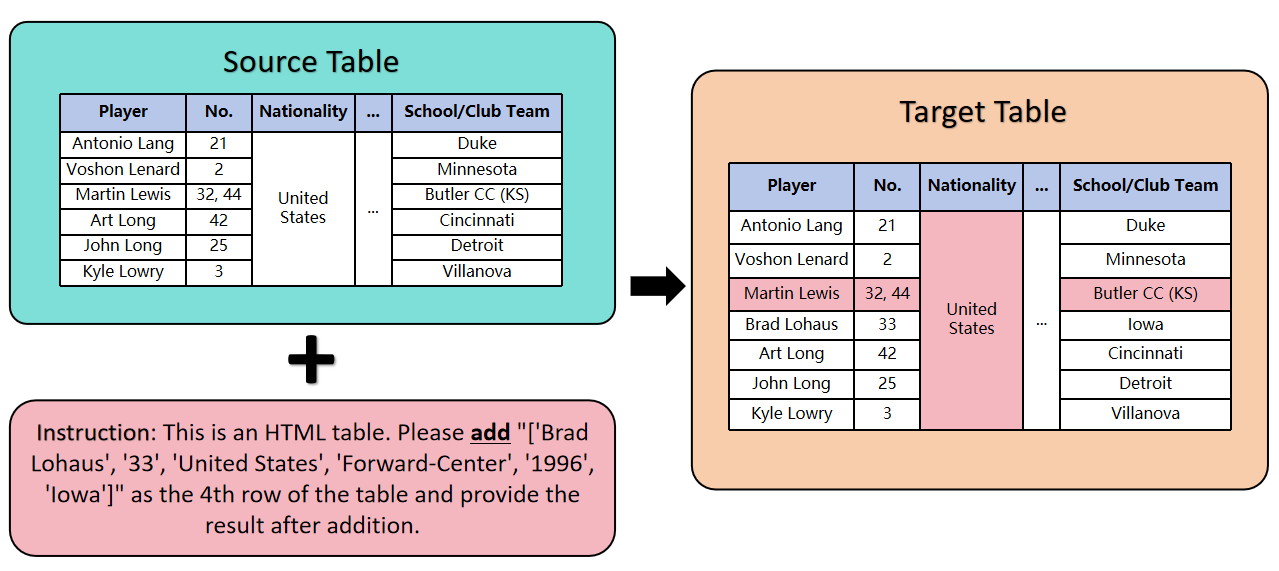}}
\subfloat[Removing a row from the regular table.]{
		\includegraphics[scale=0.33]{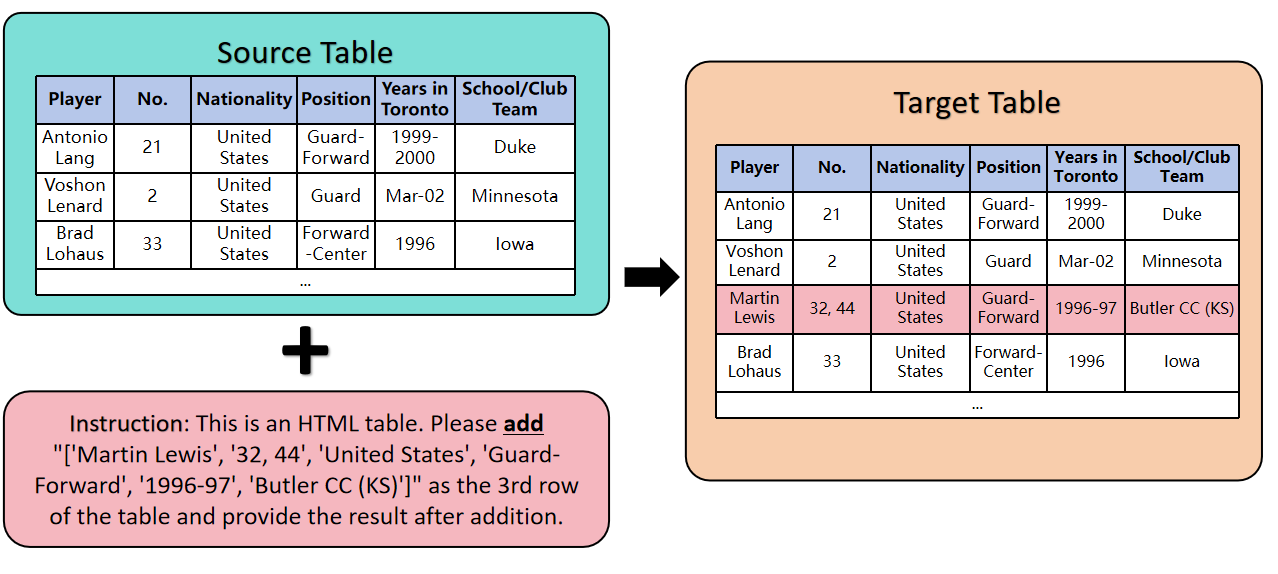}}
\\
\subfloat[Removing a column from the regular table.]{
		\includegraphics[scale=0.33]{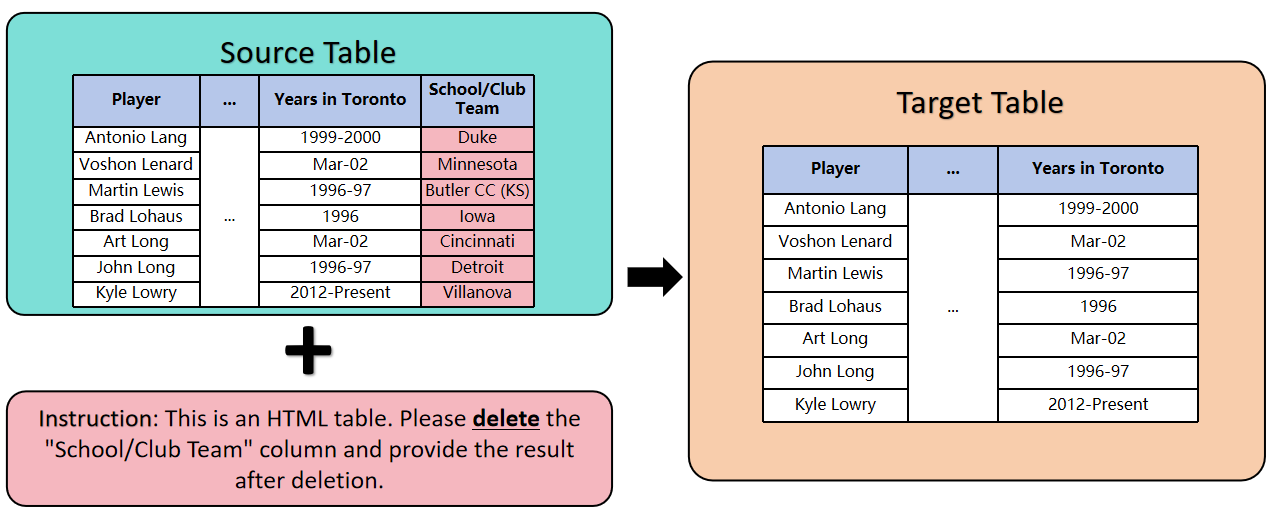}}
\subfloat[Removing a row from the irregular table.]{
		\includegraphics[scale=0.33]{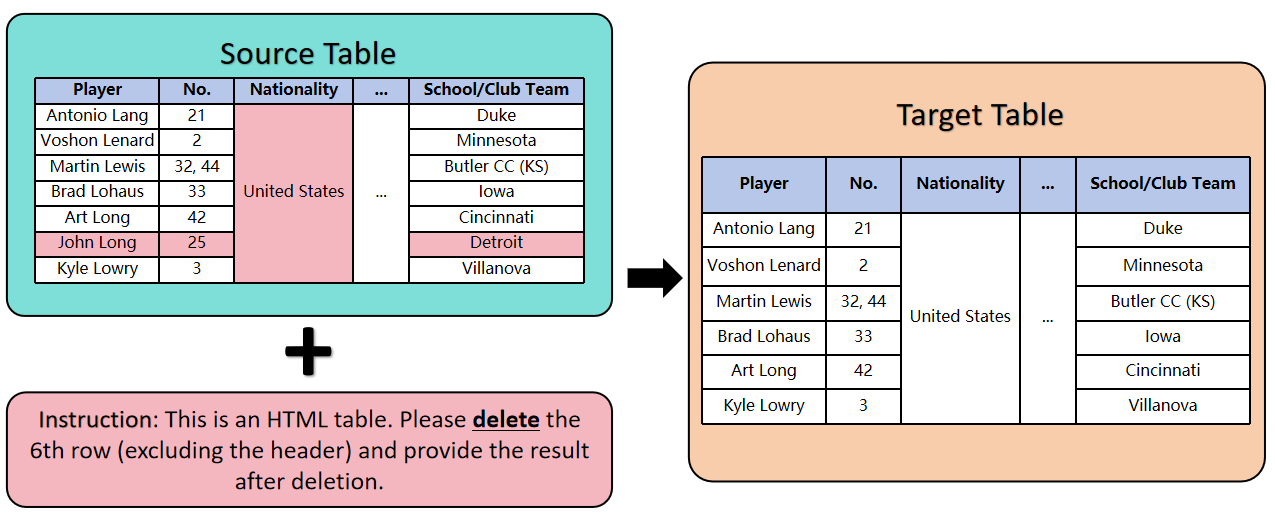}}
\\
\subfloat[Swapping two rows in the regular table.]{
		\includegraphics[scale=0.33]{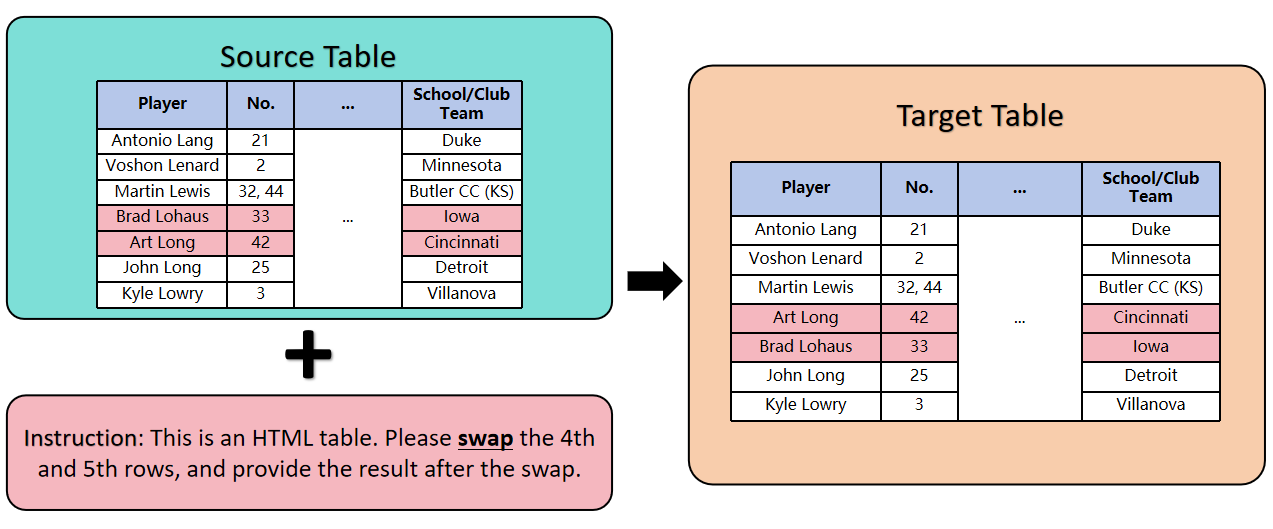}}
\subfloat[Swapping two rows in the irregular table.]{
		\includegraphics[scale=0.33]{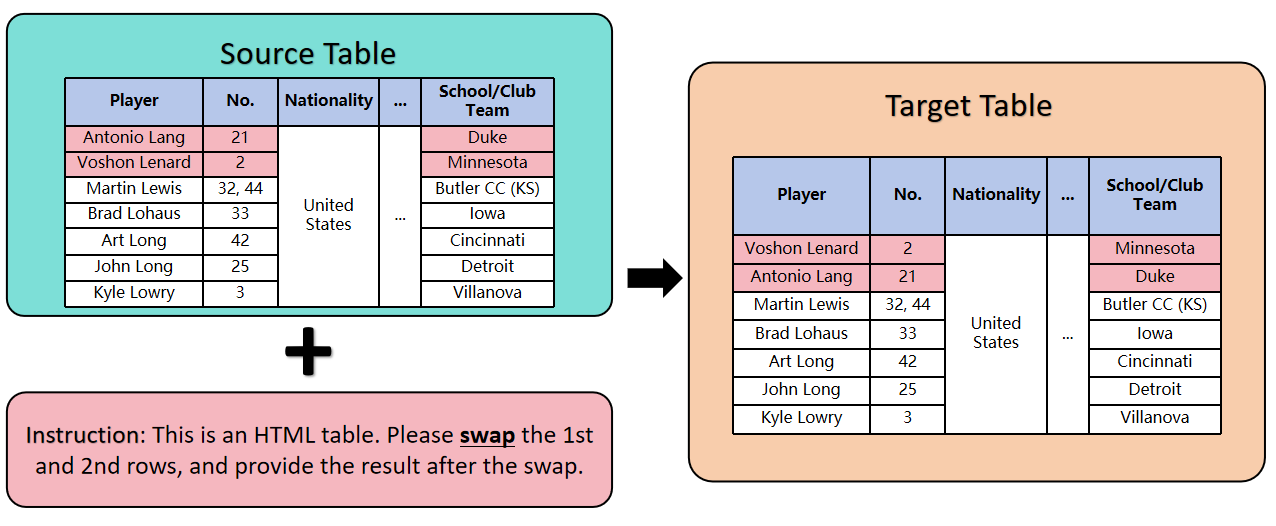}}
\\
\subfloat[Reordering the regular table based on the values in a specific column.]{
		\includegraphics[scale=0.33]{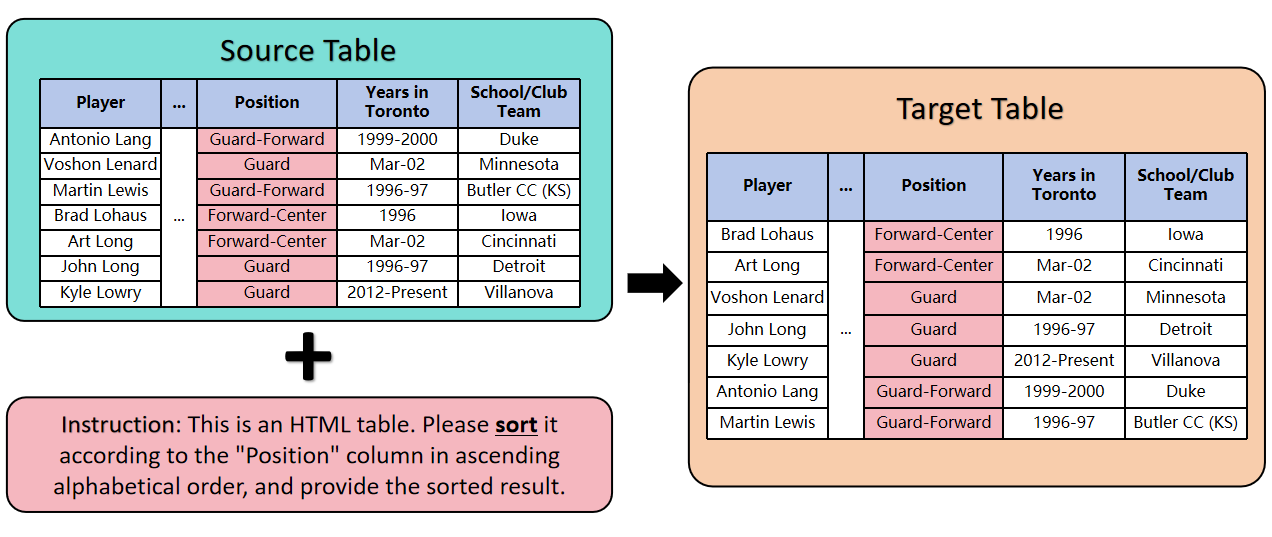}}
\subfloat[Reordering the irregular table based on the values in a specific column.]{
		\includegraphics[scale=0.33]{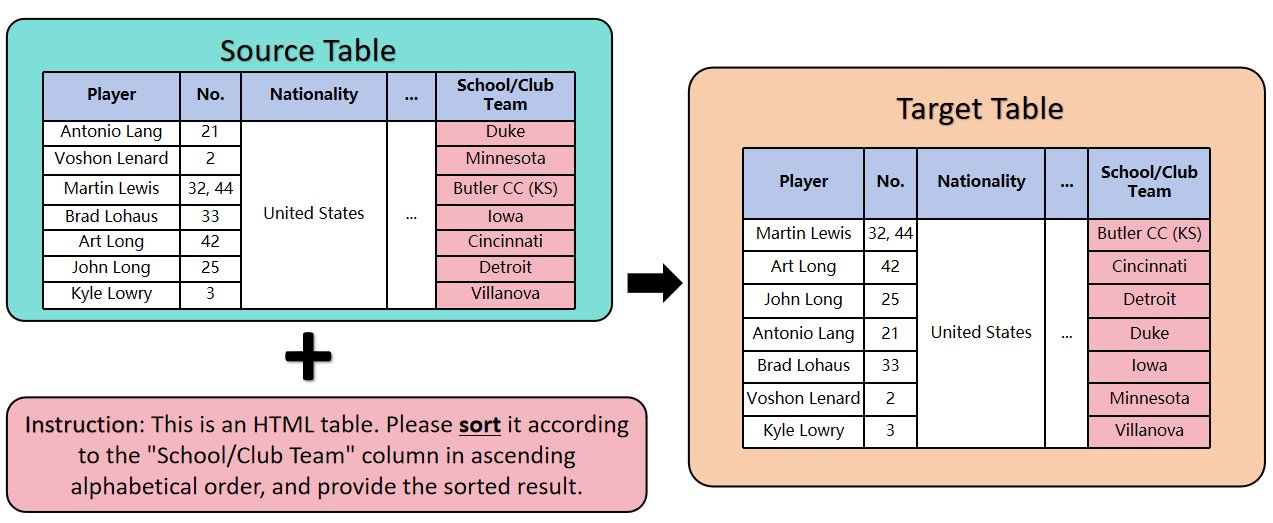}}
\\
\subfloat[Merging all adjacent identical cells in a column.]{
		\includegraphics[scale=0.33]{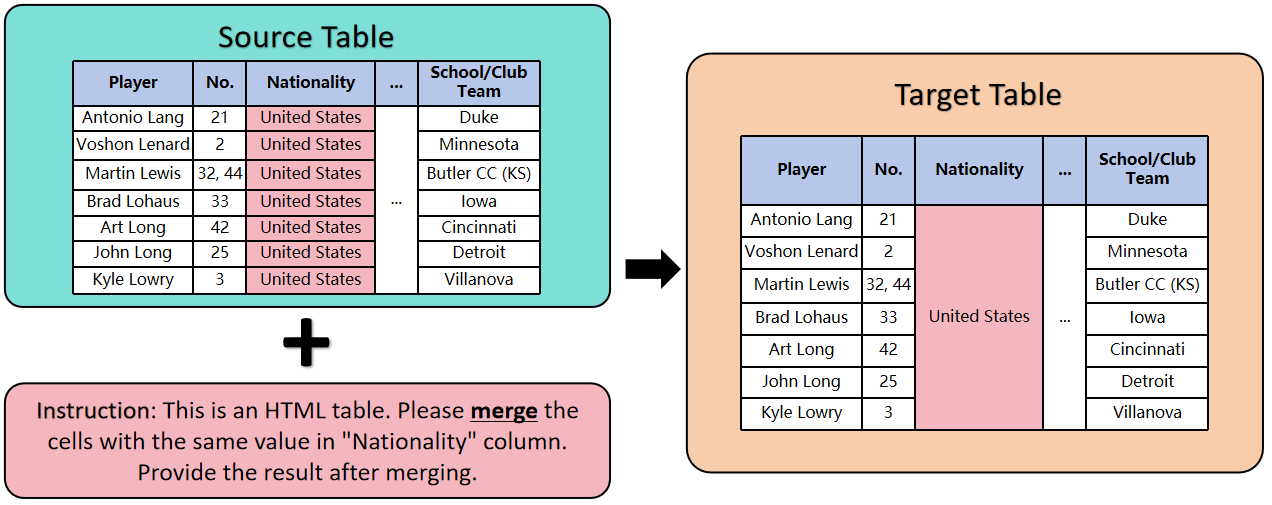}}
\subfloat[Splitting all merged cells in the table.]{
		\includegraphics[scale=0.33]{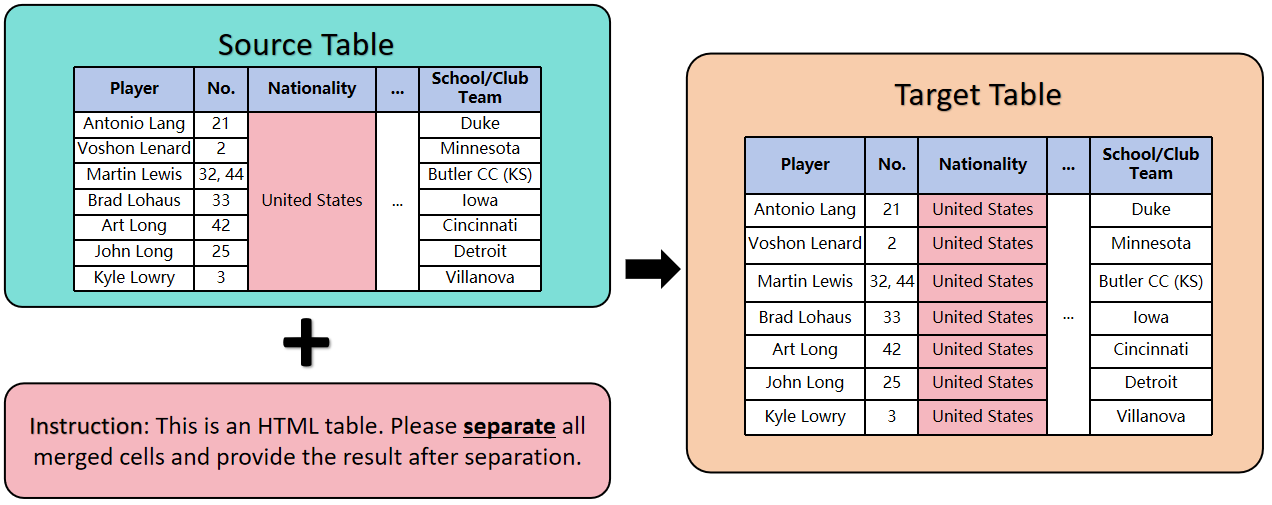}}
\caption{Examples for WikiTableEdit}
\label{examples}
\end{figure*}

To facilitate understanding of WikiTableEdit, we provide some simple examples for each operation along with its various operation targets. The examples are shown in the Figure ~\ref{examples}.

For better visibility of the modified sections, we use a pink background for highlighting. Since the Reordering operation causes a complete change in the table, we only highlight the column used as the key value. Additionally, ellipses (...) in the table indicate several cells and do not include merged cells.

\section{Results of supervised fine-tuning}

\begin{table}[!h]
\centering
\small
\setlength{\tabcolsep}{3mm}{
\begin{tabular}{l|ccc|cc}
\toprule
\textbf{Model}           & \textbf{EM} & \textbf{SARI} & \textbf{TED} & \textbf{BLEU} & \textbf{ROUGE-L}\\
 \midrule
 LLaMA2-7B & 0 & 12.02 & 15.49 & 1.32 & 28.39  \\

 ChatGLM3-6B & 0 & 23.30  & 33.06 &  17.83 & 53.35   \\
 Falcon-7B & 0  & 3.74  & 0.34 &  0.18 & 9.02   \\

 GPT-3.5-turbo & 6.34 & 54.21  & 53.81 & 13.72  & 65.74   \\
\bottomrule

\end{tabular}}
\caption{Performance of different models after fine-tuning on the WikiTableEdit test set. Please note that due to cost considerations, we do not fine-tune ChatGPT. The table provides the zero-shot performance of ChatGPT as a reference.}
\label{tab:intent}
\end{table}

\end{document}